\documentclass[10pt]{article}
\usepackage{tommoro-report}
\usepackage{wrapfig}
\usepackage{float}
\usepackage{placeins}
\usepackage{kotex}
\usepackage{amssymb}
\usepackage{amsmath}
\usepackage{graphicx}
\usepackage{booktabs}
\usepackage{multirow}
\usepackage{xcolor}
\usepackage{cite}
\usepackage{tikz}
\usetikzlibrary{arrows.meta,positioning,calc,fit}

\newcommand{\gcvla}{GC-VLA}

\settrtitle{HABILIS Brain 0: Geometry-Change Supervision for Vision-Language-Action and Residual Flow Recovery}

\begin{document}

\begin{trabstract}
Vision-language-action policies benefit from geometric supervision, but
current-frame geometry alone does not explicitly describe the changes
associated with manipulation. This design is motivated by the goal of
learning an embodiment-agnostic visual interface that can be pretrained
across robot and egocentric video before robot-specific action alignment.
We introduce Geometry-Change VLA (GC-VLA), which learns to predict
multiview future--current geometry-change tokens from current observations. Offline frame pairs define a nominal
0.5-second prediction horizon; future observations are used only to
construct training targets. Stage~1 trains a geometry-change
vision-language model (GC-VLM). Stage~2 introduces a continuous
ActionExpert and aligns it with robot actions while stopping action-flow
gradients at the VLM interface. Stage~3 enables these gradients to update
the trainable VLM components jointly with the ActionExpert. Stage~4
freezes GC-VLA and applies Geometry-Conditioned Residual Flow (GCRF),
using a binary intervention router and a single bounded residual
velocity policy learned from closed-loop feedback. GC-VLA achieves
95.20\% success on LIBERO, and GC-VLA with GCRF achieves 99.55\%.
Inference uses current observations and the learned GC representation
without executing the offline target encoders.
\end{trabstract}

\section{Introduction}

Robot manipulation requires reasoning about not only what is currently
visible, but also what should change as an action is executed. Static
current-depth supervision can provide useful scene geometry, yet it does not
directly describe the geometric transition associated with an action horizon.
For manipulation, such transitions include a gripper approaching an object, a
drawer moving along its constraint, contact becoming established, or an object
changing pose toward a task goal.

Recent world-model and predictive-representation approaches address temporal
reasoning by predicting future observations, future latent states, or
temporally evolved scene representations~\cite{videovla,flare,ahead}.
VideoVLA jointly generates future visual outcomes and robot actions, while
FLARE aligns learned future tokens with latent representations of future
observations and AHEAD explicitly rolls predicted future VLA features forward
for downstream action decoding. These methods motivate future prediction as a
useful learning signal for robot control, but leave open the question of
\emph{what} future information should be represented. GC-VLM takes a
deliberately different target. Rather than reconstructing a future observation
or predicting the complete future latent state, it predicts the change between
spatial representations of the current and future observations:
\begin{equation}
\Delta E_{t,H}=E_{t+H}-E_t.
\end{equation}
The future observation is used only to construct this supervision target
during training. The resulting objective emphasizes spatial locations and
features that change over the manipulation horizon while reducing the
contribution of unchanged scene content. At deployment, no future frame is
observed and no learned world model is rolled forward; GC-VLM predicts the
geometry-change representation directly from the current observation and
instruction, and this representation conditions action generation.

This formulation also provides an interface for heterogeneous pretraining.
Robot-action supervision is unavailable for human video and is inconsistent
across robot embodiments, action spaces, and data sources. In contrast,
temporally paired visual observations are available much more broadly.
GC-VLM therefore learns camera-aligned geometry-change targets before requiring
a common robot-action representation. This allows geometry pretraining to use
both robot demonstrations and action-free video, including egocentric human
interaction data, while postponing embodiment-specific action conversion to
subsequent robot-action alignment.

We instantiate this idea as a four-stage pipeline. Stage1 trains GC-VLM with
multiview future–current geometry-change supervision without requiring robot
actions. Stage2 connects the learned representation to a continuous
ActionExpert while stopping the action-flow gradient at the VLM interface.
Stage3 removes this detach boundary and jointly adapts the trainable VLM
components and ActionExpert on downstream robot demonstrations. Finally,
Stage4 freezes GC-VLA and applies Geometry-Conditioned Residual Flow (GCRF),
which learns selective bounded corrections to the action-generation flow from
closed-loop outcomes.

At inference, GC-VLA receives only the current language instruction, RGB
observations, and robot state. Depth Anything v3~\cite{depthanything3} and DINOv2~\cite{dino} are used exclusively for offline
construction of geometry-change supervision; neither model, future
observations, nor future-derived targets are available to the deployed policy.

The paper documents a four-stage pipeline and the resulting evaluations:
\begin{itemize}
    \item \textbf{GC-VLM geometry pretraining.}
    Multiview future--current geometry targets support representation
    learning without requiring robot-action annotations.
    \item \textbf{GC-VLA action alignment and adaptation.}
    Detached action alignment is followed by coupled optimization of
    the ActionExpert and trainable VLM components.
    \item \textbf{GCRF residual post-training.}
    A binary router and one bounded residual policy adapt the frozen
    action-generation flow using closed-loop feedback.
    \item \textbf{LIBERO evaluation.}
    GC-VLA achieves 95.20\% success, and GC-VLA with GCRF achieves
    99.55\% under the reported evaluation protocol.
\end{itemize}

\section{Related Work}

\textbf{Vision-language-action policies.}
Large-scale VLA policies establish the setting in which a vision-language backbone is adapted to robot actions. RT-1 and RT-2 scale transformer policies and vision-language-action transfer for robot control~\cite{rt1,rt2}; OpenVLA provides an open generalist VLA trained on diverse robot demonstrations~\cite{openvla}. Most directly, $\pi_0$ couples a pretrained VLM to a continuous action expert trained with flow matching~\cite{pi0}. GC-VLA follows this VLM-to-continuous-action design and uses
geometry change over a fixed prediction horizon as VLM-side supervision.

\textbf{Geometry-aware and future-state supervision.}
Geometry-aware objectives and future-state prediction provide
task-relevant structure for robot policies~\cite{gam,dp3}.
GC-VLM uses future--current geometry change as a supervision target.
At inference, action generation is conditioned on GC representations
predicted from current observations.

\textbf{Action diffusion and flow matching.}
Diffusion Policy models robot action chunks through conditional denoising~\cite{diffusionpolicy}. Flow Matching provides the general vector-field learning objective used to integrate a continuous flow from noise to data~\cite{flowmatching}; FlowPolicy further applies consistency flow matching to manipulation policies~\cite{flowpolicy}. Our GC-VLA action expert uses this continuous-action viewpoint, and Stage~4 applies corrections in flow-velocity space rather than directly to the final action chunk.

\textbf{Residual policy adaptation.}
Residual Reinforcement Learning augments a controller with an
action-space residual~\cite{residualrl}, while Recovery RL learns
selective intervention for safe execution~\cite{recoveryrl}.
GCRF applies a bounded residual to the action-generation velocity field.
Guided flow methods likewise modify the vector field during
sampling~\cite{guidedflows}, and Residual Flow Steering studies
adaptation of frozen flow policies~\cite{rfs}.

\section{Method}

\subsection{Geometry-Change Targets}

Let the online input be
\begin{equation}
    x_t=\{\ell, I_t^{\mathrm{head}}, I_t^{\mathrm{wrist-L}},
    I_t^{\mathrm{wrist-R}}, q_t\},
\end{equation}
where $\ell$ is the language instruction, $I_t^{(v)}$ is an available RGB
view, and $q_t$ is the robot state. An offline geometry encoder $\phi$ maps a
camera observation and its pseudo-depth estimate to a compact spatial
representation:
\begin{equation}
    E_t^{(v)} = \phi(I_t^{(v)}), \quad
    E_{t+H_{\mathrm{GC}}}^{(v)} = \phi(I_{t+H_{\mathrm{GC}}}^{(v)}).
\end{equation}
The training label is the geometry change over the prediction horizon
\begin{equation}
    \Delta E_{t,H_{\mathrm{GC}}}^{(v)} = E_{t+H_{\mathrm{GC}}}^{(v)}-E_t^{(v)}
    \in\mathbb{R}^{10\times10\times d}.
\end{equation}
The $10\times10$ spatial grid is quantized into 100 token positions per view,
$y_{t,H_{\mathrm{GC}}}^{(v)}\in\{1,\ldots,K\}^{100}$. The fixed-layout target is
\begin{equation}
    y_{t,H}^{\mathrm{GC}}=
    [y_{t,H}^{\mathrm{head}};
     y_{t,H}^{\mathrm{wrist-L}};
     y_{t,H}^{\mathrm{wrist-R}}]\in\{1,\ldots,K\}^{300}.
\end{equation}

For source frame rate $f$, the target offset is
$H_{\mathrm{GC}}=\operatorname{round}(0.5f)$ frames.
The action chunk length, denoted $H_A$, is defined separately when
robot-action supervision is introduced. Missing cameras retain their
assigned token slots and are excluded by a validity mask.
The head view captures scene-level change, while wrist views provide
local observations of gripper--object interaction.

Robot state $q_t$ is used by the action policy; geometry pretraining
uses the available visual and language inputs without requiring robot
state or action annotations.

\subsection{Architecture}

\begin{figure}[t]
    \centering
    \includegraphics[width=0.98\textwidth]{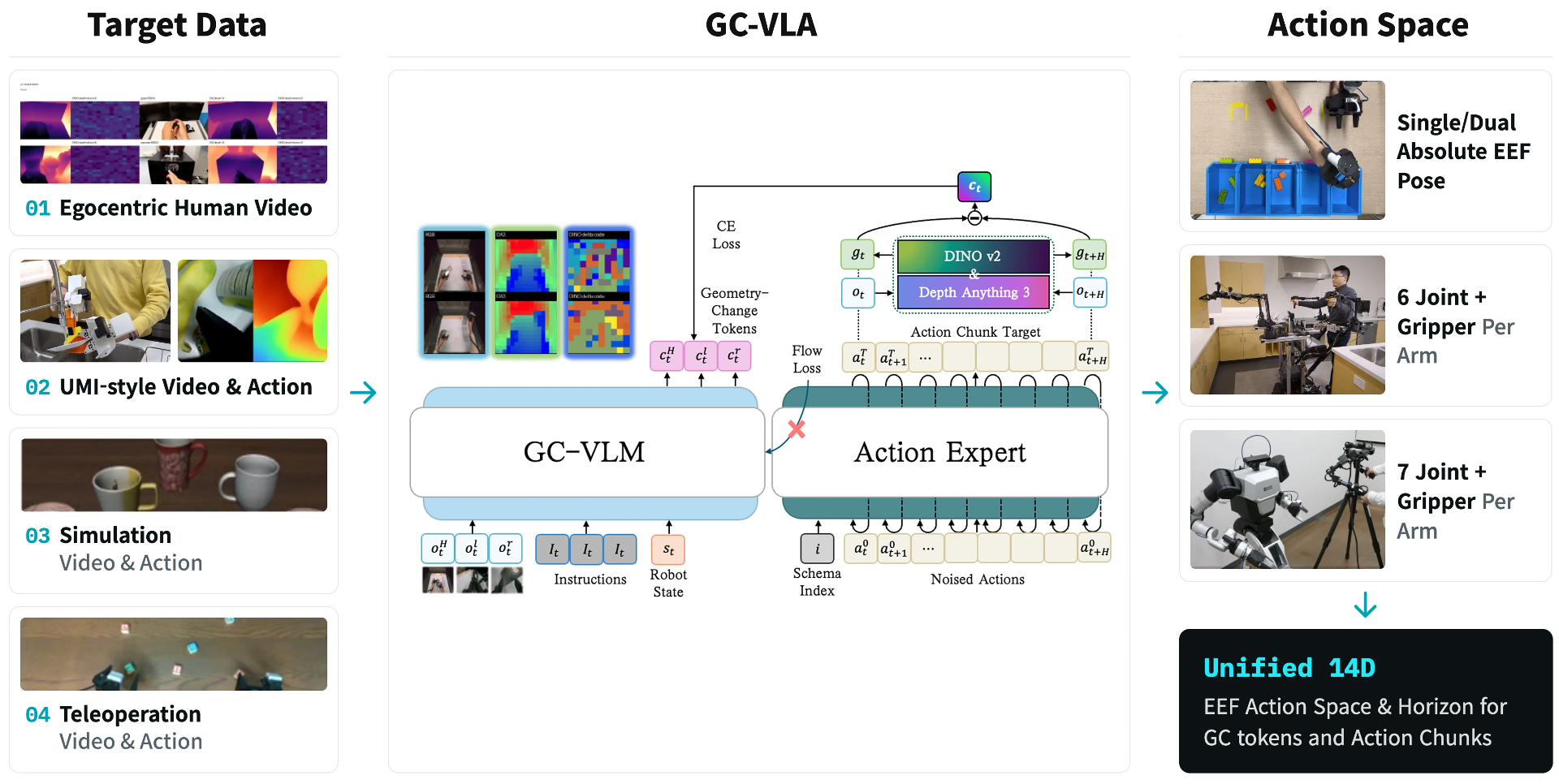}
    \caption{GC-VLA architecture and training stages. GC-VLM predicts
multiview geometry-change tokens from current observations. The
ActionExpert is introduced during action alignment and consumes the
learned representations. The action-flow gradient is detached at the
VLM interface in Stage~2 and enabled for trainable VLM components in
Stage~3. Future observations are used only for offline target generation.}
    \label{fig:gcvla_architecture}
\end{figure}

GC-VLM extends the 36-layer Molmo2-ER backbone with 12
additional transformer blocks, initialized by copying the
first 12 pretrained blocks. The appended blocks form the upper
geometry-change extension and are trained to support prediction of
multiview geometry-change tokens. Stage~1 learns this representation
without action supervision. Stage~2 trains a continuous ActionExpert
conditioned on the learned visual-language and GC representations.
Figure~\ref{fig:gcvla_architecture} summarizes the architecture.

The shared VLM first produces multimodal hidden states
\begin{equation}
    h_t=F_{\Theta}(x_t), \qquad
    \hat y_{t,H}^{\mathrm{GC}}=g_{\mathrm{GC}}(h_t).
\end{equation}
Geometry-change prediction therefore shapes the representation consumed by the
action branch. In addition, the implemented \emph{GeometryReader} pools the GC
hidden states with learned queries and exposes them to selected ActionExpert
blocks through cross-attention residual updates:
\begin{equation}
    \tilde h_{t,b}^{A}=h_{t,b}^{A}+
    \operatorname{Reader}_{b}(h_{t,b}^{A},h_t^{\mathrm{GC}},m_t^{\mathrm{GC}}),
    \qquad b\in\mathcal B_{\mathrm{GC}}.
\end{equation}
Here $m_t^{\mathrm{GC}}$ masks unavailable camera slots. The continuous
ActionExpert then predicts the flow velocity conditioned on the resulting
geometry-aware hidden states,
\begin{equation}
    v_{k}^{\mathrm{base}}=
    A_{\Psi}(z_k,\tau_k,q_t;h_t,\tilde h_t^{A}).
\end{equation}
The reader is initialized as a no-op through a zero-initialized output
projection, preserving the pretrained action path at initialization while
allowing geometry-conditioned updates to be learned.

\subsection{Leakage Prevention}

Because $\Delta E_{t,H}$ is built from $I_{t+H}$, future information must never enter the evaluation path.
All rollout experiments use only $x_t$ at inference.
The intended invariant is
\begin{equation}
    \texttt{eval: } \hat y_{t,H}^{\mathrm{GC}}=g_{\mathrm{GC}}(F_\Theta(x_t)),
    \quad \text{not } \phi(I_{t+H}).
\end{equation}
Future frames are labels, not inputs.

\subsection{GCRF: Geometry-Conditioned Residual Flow}
GCRF adapts a frozen GC-VLA policy using one binary intervention router
and one unified residual velocity policy. The base VLM and ActionExpert
parameters remain fixed during this stage.

At replanning instant $t$, the residual policy receives a context $c_t$
formed from base-policy, GC-conditioned, and flow-history features.
\begin{equation}
    c_t=[\bar v_t^{\mathrm{base}};g_t^{\mathrm{GC}};
         h_t^{\mathrm{flow}}]\in\mathbb{R}^{96},
\end{equation}
where each component is 32-dimensional. The GC component is computed
from the model's intermediate representations rather than an online
execution of the offline target encoders.
Let $b_t\in\{0,1\}$ denote the router decision and let $z_t$ denote
the selected residual latent. The bounded correction is
\begin{equation}
    d_t=\epsilon\tanh(z_t).
\end{equation}
The base action flow is integrated as
\begin{equation}
    x_{t,k+1}=x_{t,k}+\Delta\tau_k
    \left[v_{\mathrm{base}}(x_{t,k},\tau_k\mid o_t)+b_t d_t\right],
    \qquad k=0,\ldots,9.
\end{equation}
The correction is held constant across the ten integration steps of
one replanning cycle. A subsequent replan can produce a different
correction as the observation and context change. At every integration
step, the base velocity is evaluated at the updated flow state.

The router is learned from the observed outcomes of base-policy and
intervention rollouts. At inference it uses observation-derived
features; success labels are used only during training. When $b_t=0$,
no residual velocity is added. When $b_t=1$, the selected correction
is applied within the base flow solver.

\section{Experimental Setup}

\subsection{Pretrained Geometry Models and Target Generation}
\begin{figure*}[t]
    \centering
    \includegraphics[width=\textwidth]{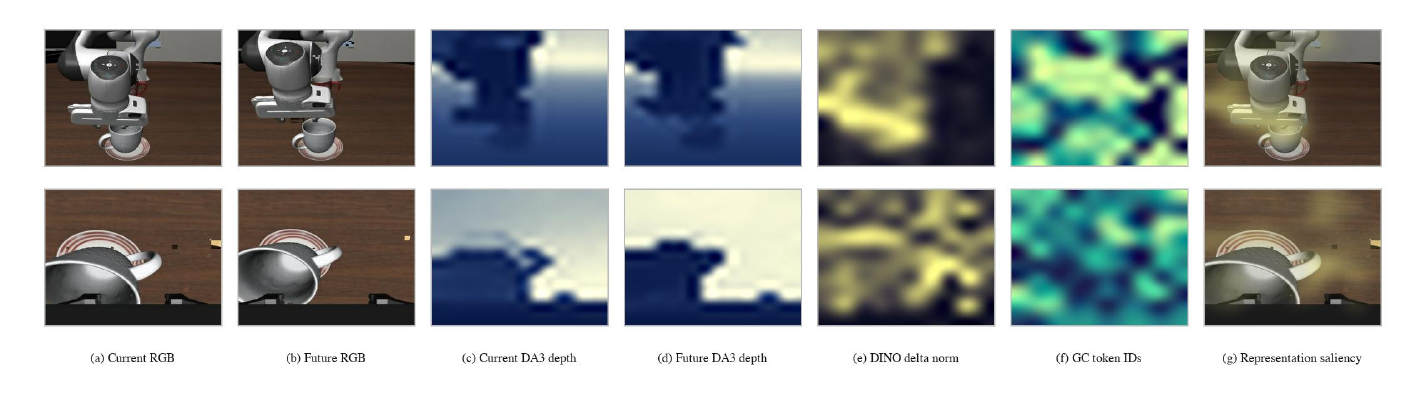}
    \caption{Multiview geometry-change target construction on LIBERO.
Rows correspond to head and wrist views. Columns show (a) current RGB,
(b) future RGB at the selected geometry horizon, (c--d) their offline
DA3 pseudo-depth estimates, (e) the magnitude of the DINO feature
difference, (f) the quantized GC token IDs, and (g) the GC target
overlay on the current RGB frame. The overlay visualizes the training
target rather than model attention. Future observations are used only
to construct offline supervision.}
    \label{fig:gc_multiview_representation_actual}
\end{figure*}

We use two frozen pretrained models only during offline target generation. For pseudo-depth estimation, we use \texttt{depth-anything/DA3-BASE}, a 0.12B-parameter checkpoint from Depth Anything 3~\cite{depthanything3}. For geometry representation extraction, we use \texttt{facebook/dinov2-small}, corresponding to the 21M-parameter distilled DINOv2 ViT-S/14 model~\cite{dino}. Both checkpoints are used as preprocessing models and are not executed during GC-VLA inference. For each camera view, Depth Anything 3 first predicts a pseudo-depth map from the current RGB observation. The depth map is resized to a $16\times16$ grid and normalized using per-frame percentile normalization. The resulting single-channel depth grid is replicated to three channels and resized to $224\times224$ before being passed to DINOv2. We remove the DINOv2 class token, resize the patch-token map to $10\times10$, and apply feature normalization. For a current frame $o_t$ and a future frame $o_{t+H}$, where $H=\mathrm{round}(f\cdot0.5)$ for a source frame rate $f$, the geometry target is computed as\begin{equation}e_t=\psi(\mathrm{DA3}(o_t)), \qquad e_{t+H}=\psi(\mathrm{DA3}(o_{t+H})).\end{equation}\begin{equation}\Delta e_{t,H}=e_{t+H}-e_t.\end{equation}The continuous DINOv2 feature difference is then quantized into 100 discrete codes per view. The canonical cache contains three fixed view slots: head/global view, left-wrist view, and right-wrist view, resulting in 300 geometry tokens per sample. Missing camera views are represented by invalid masks and are not replaced by another camera. Figure~\ref{fig:gc_multiview_representation_actual} illustrates the
target construction for head and wrist observations.

\paragraph{Pretrained model licensing.}The DINOv2 code and model weights are released under Apache License 2.0. The DA3-BASE model card also lists the checkpoint under Apache 2.0. We use both models only for offline geometry-target generation and retain the corresponding attribution and license notices in the released implementation. GC-VLM initialization starts from a Molmo2-ER vision-language checkpoint. 

\subsection{Stage-wise Training Data and Optimization}

Stages~1--3 use distinct data mixtures with compatible visual and
geometry-target interfaces. Their action supervision and gradient
routing differ as follows.

\paragraph{Stage~1: GC-VLM geometry pretraining.}
GC-VLM is initialized from Molmo2-ER and trained on current--future
frame pairs with offline geometry-change targets~\cite{molmoact2}. Robot-action labels
are not required, and no ActionExpert is used in this stage.

\paragraph{Stage~2: Detached action alignment.}
A continuous ActionExpert is introduced and trained on robot
demonstrations converted to a common 14-dimensional bimanual
end-effector delta representation. Each arm contributes three
translation components, three rotation components, and a gripper
command. Unavailable arm dimensions are masked. Geometry supervision
continues to train GC-VLM, while action-flow gradients are stopped at
the VLM--ActionExpert interface.

\paragraph{Stage~3: Coupled VLA adaptation.}
The model is adapted to LIBERO with geometry and action supervision.
The detach boundary is removed so that action-flow gradients update
the trainable VLM components together with the ActionExpert.

Let $\mathcal L_{\mathrm{CE},s}$ denote the stage-specific token
cross-entropy and $\mathcal L_{\mathrm{FM}}$ the action flow-matching
loss. The gradient-routing curriculum is
\begin{align}
    \mathcal L_1 &= \mathcal L_{\mathrm{GC}},
       &&\text{no ActionExpert},\\
    \mathcal L_2 &= \mathcal L_{\mathrm{CE},2}
       +\lambda_{\mathrm{FM}}\mathcal L_{\mathrm{FM}},
       &&\left.\frac{\partial\mathcal L_{\mathrm{FM}}}
                       {\partial h_t}\right|_{\mathrm{interface}}=0,\\
    \mathcal L_3 &= \mathcal L_{\mathrm{CE},3}
       +\lambda_{\mathrm{FM}}\mathcal L_{\mathrm{FM}},
       &&\text{action-flow gradients enabled}.
\end{align}
The token objectives include the valid geometry targets and any
action-token targets enabled by the corresponding training recipe.

\paragraph{Stage~4: Residual post-training.}
GC-VLA is frozen. Closed-loop feedback is used to train selective
residual intervention, as described in
Section~\ref{sec:gcrf-training}.

\subsection{Unified End-effector Action Representation}
Source demonstrations are converted to a common bimanual end-effector
delta representation before action training. For arm $u\in\{L,R\}$,
\begin{equation}
    a_t^u=[\Delta p_t^u;\Delta\rho_t^u;g_{t,\mathrm{cmd}}^u]
       \in\mathbb{R}^{7},\qquad
    a_t=[a_t^L;a_t^R]\in\mathbb{R}^{14}.
\end{equation}
Here $\Delta p$ denotes translation, $\Delta\rho$ denotes the
three-component rotation increment under the implemented conversion
convention, and $g_{\mathrm{cmd}}$ is the gripper command rather than
a gripper-state difference. Left and right arm blocks have fixed
positions. Missing arm dimensions are masked.

Each source adapter resolves pose conventions, units, arm ordering,
gripper semantics, and action timing before conversion. Absolute EEF
poses are converted into adjacent transitions. Joint-position sources
require forward kinematics using the corresponding robot model and
frame calibration before EEF conversion. Sources without a validated
conversion are excluded from action training.

The model may carry these values in a padded action tensor; padding
does not add physical action dimensions. Invalid dimensions are
excluded from the action loss.

\subsection{Data}

The canonical GC-VLA lineage uses distinct data mixtures at each stage.
Stage~1 pretrains the geometry-change representation on current--future frame
pairs drawn from diverse human and robot video sources. Stage~2 uses
physically validated action-bearing robot caches, with source-specific action
schemas mapped to the common masked 32-dimensional transport interface.
LIBERO is excluded from these broad
pretraining mixtures: it is introduced only for downstream Stage~3 adaptation
and is retained as the canonical closed-loop evaluation benchmark.

\subsection{Training Data}
\begin{figure*}[t]
    \centering
    \includegraphics[width=\textwidth]{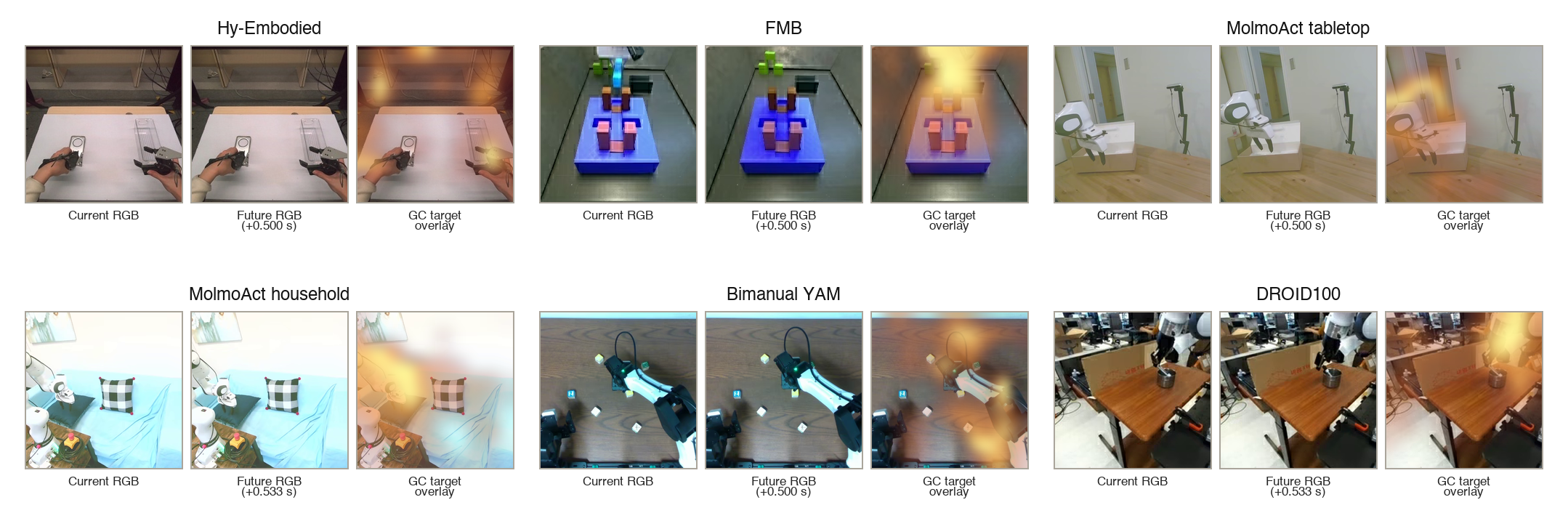}
    \caption{Examples of geometry-change supervision across training
    sources. Each example contains current RGB, future RGB at the
    source-normalized prediction horizon, and a GC target overlay on
    the current image. The overlay is the spatial magnitude of the
    DA3--DINO feature difference before quantization. Frame selection
    uses $H_{\mathrm{GC}}=\operatorname{round}(0.5f)$ for source frame
    rate $f$; the realized interval depends on frame discretization.
    Future RGB is used only to construct training targets.}
    \label{fig:gc-training-data-diversity}
\end{figure*}

Table~\ref{tab:stage-data} lists the source families used at each stage.
Stage~1 uses human and robot videos for geometry supervision without
requiring action annotations. Stage~2 uses robot demonstrations with
validated EEF action conversion. LIBERO is introduced during Stage~3
adaptation. Stage~4 uses closed-loop rollouts of the frozen GC-VLA
policy and residual candidates.
Figure~\ref{fig:gc-training-data-diversity} shows representative
current--future pairs and GC target overlays from the training sources.

\begin{table}[t]
\centering
\footnotesize
\setlength{\tabcolsep}{3pt}
\renewcommand{\arraystretch}{1.08}
\caption{Training-data sources by stage. Stage~1 uses geometry targets;
Stage~2 uses validated EEF action targets; Stage~3 uses LIBERO for
coupled adaptation. Stage~4 uses closed-loop rollouts.}
\label{tab:stage-data}
\begin{tabular}{p{0.27\columnwidth}p{0.07\columnwidth}p{0.07\columnwidth}p{0.07\columnwidth}p{0.40\columnwidth}}
\toprule
Source family & S1 & S2 & S3 & Primary role \\
\midrule
HoloAssist~\cite{holoassist2023} & Y & -- & -- & Egocentric human-interaction geometry; no robot action loss \\
Hy-Embodied~\cite{hyembodied2026} & Y & Y & -- & Cross-embodiment and bimanual geometry/action alignment \\
HABIT~\cite{habit2026} & Y & Y & -- & Human-present bimanual and shared-workspace manipulation \\
MolmoAct tabletop/household~\cite{molmoact2025,molmoact2} & Y & Y & -- & Language-conditioned tabletop and household manipulation \\
Bimanual YAM~\cite{molmoact2} & Y & Y & -- & Top and dual-wrist bimanual data; FK-normalized EEF actions in S2 \\
FMB / DROID100~\cite{fmb2024,droid2024} & Y & Y & -- & Contact-rich and real-world single-arm manipulation \\
ALOHA~\cite{aloha2023} & Y & -- & -- & Dual-wrist bimanual geometry pretraining \\
Austin Sailor / Berkeley UR5~\cite{openxembodiment2024}
& -- & Y & -- & Single-arm demonstrations converted to EEF actions \\
Berkeley Fanuc~\cite{openxembodiment2024}
& -- & Y & -- & Single-arm demonstrations converted to EEF actions \\
LIBERO~\cite{libero} & -- & -- & Y & Downstream coupled adaptation and closed-loop evaluation \\
\bottomrule
\end{tabular}

\vspace{1pt}
{\footnotesize Stage~2 includes sources with validated EEF conversion,
camera-slot mapping, and temporal alignment. Unavailable action and
camera dimensions are masked.}
\end{table}

\subsection{Action Conversion and Temporal Alignment}
Action preprocessing decodes each source format, transforms it into
the canonical EEF frame convention, normalizes translation and
gripper units, and constructs adjacent EEF transitions. Joint-state
sources undergo validated forward kinematics before this conversion.
The resulting left and right arm targets are concatenated and masked
as specified in the EEF representation above.

An action chunk comprises consecutive transitions,
\begin{equation}
    A_t=[a_t,a_{t+1},\ldots,a_{t+H_A-1}],
\end{equation}
rather than displacements measured from the initial chunk state.
Action and geometry supervision are aligned using each source's
validated timestamps and sampling rate. At canonical LIBERO
inference, the policy predicts a horizon-10 action chunk.

\subsection{Optimization Summary}
Table~\ref{tab:training-stages} summarizes the stages and their
gradient-routing rules. Geometry-token losses mask unavailable view
slots, and action losses mask unavailable action dimensions.
The Stage~2 interface is detached for the flow objective; Stage~3
enables action-flow gradients through the trainable VLM components.

\begin{table}[t]
\centering
\footnotesize
\setlength{\tabcolsep}{3pt}
\renewcommand{\arraystretch}{1.08}
\caption{Four-stage training recipe. ``Detach'' stops the action-flow gradient at the VLM--ActionExpert interface.}
\label{tab:training-stages}
\begin{tabular}{p{0.14\columnwidth}p{0.43\columnwidth}p{0.31\columnwidth}}
\toprule
Stage & Data and target & Learning rule \\
\midrule
1: GC-VLM & Video pairs and geometry-change targets
& Geometry-token CE; no ActionExpert \\
2: detached GC-VLA & Robot demonstrations with unified EEF targets
& Action flow; VLM interface detached for flow gradients \\
3: coupled GC-VLA & LIBERO geometry and action supervision
& Coupled updates of trainable VLM and ActionExpert \\
4: GCRF & Closed-loop contexts, interventions, and outcomes
& Frozen base; residual and router post-training \\
\bottomrule
\end{tabular}
\end{table}

\subsection{Stage~4: GCRF---Geometry-Conditioned Residual Flow}
\label{sec:gcrf-training}
The Stage~3 GC-VLA backbone and ActionExpert are frozen.
Closed-loop executions provide contexts, sampled residual latents,
and terminal binary success labels. The residual policy is optimized
together with a continuous value head using an explicit no-op latent.

Let $Q_\omega(c,z)$ denote a success logit and let $z=0$ denote no
intervention. The value head is trained with binary cross-entropy on
observed outcomes. Successful samples additionally impose a margin
between the sampled residual and the no-op value. Policy optimization
uses centered value differences $Q_\omega(c,z)-Q_\omega(c,0)$ to weight
the log-likelihood of sampled latents, together with a term favoring
higher value at the policy mean.

During value-guided candidate selection, the policy mean is retained
when $Q_\omega(c,\mu_\theta(c))>Q_\omega(c,0)+m$; otherwise the selected
latent is zero. The binary intervention router is trained separately
from closed-loop outcome supervision.

\subsection{Evaluation Protocol and Metrics}
\label{sec:eval-protocol}
Canonical LIBERO evaluation comprises four suites with ten tasks per
suite and fifty fixed initial states per task. We report episode
success rates and the mean over suites. The reported canonical result
uses a single fixed-initialization evaluation with seed 1000 and batch
size five. The model checkpoint and evaluation configuration are
fixed for the reported run.

Representation probes report action-chunk MSE and $R^2$, and
manipulation-phase accuracy. Closed-loop success is the primary policy
metric. Diagnostic subsets and inference interventions are reported
separately from the full benchmark evaluation.
\section{Results}

\subsection{Geometry-Change Representation Probes}
We evaluate whether geometry-change representations encode
action-relevant information using matched action-chunk and
manipulation-phase probes. Table~\ref{tab:rep-probe} compares target
representations under a common split and state baseline.
Table~\ref{tab:gc-representation-probe} evaluates information retained
in frozen learned representations.

\begin{table}[t]
\centering
\footnotesize
\setlength{\tabcolsep}{3pt}
\renewcommand{\arraystretch}{1.08}
\caption{Action-chunk prediction from alternative representations
under a matched state baseline and data split.}
\label{tab:rep-probe}
\begin{tabular}{lccc}
\toprule
Representation & MSE $\downarrow$ & R$^2 \uparrow$ & MSE reduction \\
\midrule
State only & 0.1383 & 0.354 & -- \\
Current-depth VQ & 0.1327 & 0.380 & +4.1\% \\
\gcvla{} head delta-depth & 0.1256 & 0.413 & +9.2\% \\
\gcvla{} head+wrist delta-depth & \textbf{0.1064} & \textbf{0.503} & \textbf{+23.1\%} \\
\bottomrule
\end{tabular}
\end{table}

\begin{table}[t]
\centering
\footnotesize
\setlength{\tabcolsep}{3pt}
\renewcommand{\arraystretch}{1.08}
\caption{Frozen-representation linear probes under matched checkpoints,
data, and feature dimensions. Results use task-stratified 10-fold
episode cross-validation over 1,600 observations from 400 LIBERO episodes.}
\label{tab:gc-representation-probe}
\begin{tabular}{lccc}
\toprule
Probe target
& Current depth
& Geometry change
& Improvement \\
\midrule
Action-chunk $R^2$
& 0.512
& \textbf{0.532}
& $+0.020$ \\
Manipulation-phase accuracy
& 86.9\%
& \textbf{90.6\%}
& $+3.7$ pp \\
\bottomrule
\end{tabular}

\vspace{2pt}
{\footnotesize
Action-chunk gain: 95\% CI $[+0.004,+0.037]$,
one-sided sign-flip $p=0.027$.
Phase gain: positive in 9/10 folds,
Holm-adjusted $p=0.0195$.}
\end{table}

Table~\ref{tab:gc-representation-probe} complements the target-screening result in Table~\ref{tab:rep-probe}: with matched checkpoints, data, and feature dimensions, the frozen GC representation improves manipulation-phase decoding and modestly improves action-chunk prediction relative to current-depth features.

\paragraph{Target-encoder design evidence.}
Direct DA3 features retain metric-like spatial layout and were competitive in some in-domain action probes; our choice is therefore not based on a claim that direct depth is uniformly inferior. On the matched dynamic-patch benchmark, however, affine-aligned direct depth obtained AUROC/AP of $0.730/0.408$, and VQ depth obtained $0.712/0.421$, whereas the DA3--DINO feature difference obtained $0.937/0.762$. Its static-region false-positive rate was $0.027$, compared with $0.092$ for affine depth and $0.093$ for VQ depth. These measurements motivate DA3--DINO feature differences as the
target representation: the matched probe shows stronger change
discrimination and fewer static-region false positives while retaining
a spatial token grid.

For wrist observations, a pilot reconstructed camera-frame pseudo-3D trajectories from DA3 depth. It visualized local interaction motion but did not produce a contract-valid SE(3) target: monocular depth lacked stable metric scale, camera and object motion were entangled, and camera-to-robot extrinsics were unavailable or inconsistent across sources. We therefore retain a camera-aligned wrist grid and mask missing views. In a diverse-task action probe, head-only embedded current-plus-delta features achieved $R^2=0.467$, wrist-only direct current-plus-delta achieved $0.492$, and their aligned combination achieved $0.654$; shuffling wrist features reduced it to $0.379$. This supports complementary local wrist information without treating the pseudo-SE(3) pilot as a controlled policy ablation.

\subsection{Direct-Training Geometry-Change Ablation}
\label{sec:stage-free-ablation}

To isolate the effect of geometry-change supervision from the staged
pretraining curriculum, we evaluate a matched pair of policies initialized
from the same VLA checkpoint and trained directly on LIBERO without using the
Stage~1 geometry-pretraining or Stage~2 detached-alignment procedure.

The control variant uses current-depth VQ supervision with 100 geometry tokens.
The geometry-change variant replaces this target with 200 head-and-wrist
future--current depth-delta tokens while otherwise retaining the same direct
VLA training path as MolmoAct2~\cite{molmoact2}. Both models are evaluated under the same 2{,}000-episode
fixed-initialization LIBERO protocol.

\begin{table}[t]
\centering
\footnotesize
\setlength{\tabcolsep}{3pt}
\renewcommand{\arraystretch}{1.08}
\caption{Stage-free closed-loop ablation isolating the geometry target from
the Stage~1--2 pretraining curriculum.}
\label{tab:stage-free-ablation}
\begin{tabular}{lcccc}
\toprule
Policy & Geometry target & Tokens & Success & Rate \\
\midrule
Current-depth control
& Current-depth VQ & 100
& 1{,}642/2{,}000 & 82.1\% \\
Geometry-change variant
& Head+wrist depth-delta & 200
& 1{,}722/2{,}000 & \textbf{86.1\%} \\
Improvement
& -- & +100
& +80 & \textbf{+4.0 pp} \\
\bottomrule
\end{tabular}
\end{table}

Replacing current-depth supervision with geometry-change supervision improves
closed-loop success from 82.1\% to 86.1\%, corresponding to 80 additional
successful episodes, as shown in Table~\ref{tab:stage-free-ablation}. Because neither variant uses the Stage~1 or Stage~2
pretraining curriculum, this result isolates a benefit from the target design
itself rather than from the staged optimization procedure.

The subsequent 95.20\% Stage~1--3 result should therefore be interpreted as
combining two effects: the geometry-change target already improves direct VLA
training, while dedicated geometry pretraining, detached action alignment,
and coupled downstream adaptation provide additional gains.

\subsection{Stage~1--3: Canonical LIBERO Evaluation}

The Stage~1--3 GC-VLA policy achieves 95.20\% success under the
protocol in Section~\ref{sec:eval-protocol}.
Table~\ref{tab:dlf-rollout} compares the base policy with the final
GCRF configuration. Table~\ref{tab:libero-comparison} compares suite
success rates with published methods, grouped by adaptation regime.

\begin{table}[t]
\centering
\footnotesize
\setlength{\tabcolsep}{3pt}
\renewcommand{\arraystretch}{1.08}
\caption{Canonical LIBERO evaluation of the frozen GC-VLA base and the final GCRF recovery policy under the same 2{,}000-rollout contract.}
\label{tab:dlf-rollout}
\begin{tabular}{lccc}
\toprule
Training pipeline & Evaluation contract & Success & Rate \\
\midrule
\gcvla{} Stage~1--3 & 40 tasks $\times$ 50 fixed-init states & 1{,}904/2{,}000 & 95.20\% \\
\gcvla{} + GCRF & Same frozen-base evaluation cohort & \textbf{1{,}991/2{,}000} & \textbf{99.55\%} \\
Improvement & Same paired contract & +87 & +4.35 pp \\
\bottomrule
\end{tabular}
\end{table}

\begin{table*}[t]
\centering
\footnotesize
\setlength{\tabcolsep}{3pt}
\renewcommand{\arraystretch}{1.08}
\caption{Comparison on the canonical LIBERO benchmark. We report success
rates (\%) on the four standard suites---LIBERO-Spatial, LIBERO-Object,
LIBERO-Goal, and LIBERO-Long---and their unweighted mean. Methods are
grouped by adaptation regime: supervised or task-specialized policies
(upper block) and policies using closed-loop or online post-training on
the LIBERO distribution (lower block). The HABILIS Brain~0 row follows
the canonical fixed-initial-state evaluation contract described in
Section~\ref{sec:eval-protocol}. Published baseline values are reproduced
from the cited sources for contextual comparison; training data,
observation and action interfaces, reset-state cohorts,
checkpoint-selection rules, and evaluation implementations may differ
across methods. Therefore, this table should not be interpreted as a
controlled comparison of held-out-cohort generalization.
$^{\dagger}$The RLinf-GRPO average is computed over the four displayed
suites; its reported 98.1 leaderboard value additionally includes
LIBERO-90.}
\label{tab:libero-comparison}
\begin{tabular}{lccccc}
\toprule
Method & L-Spatial & L-Object & L-Goal & L-Long & Average \\
\midrule
\multicolumn{6}{l}{\textit{Supervised or task-specialized policies}} \\
ABot-M0.5~\cite{abotm05} & \textbf{100.0} & 99.8 & \textbf{99.4} & 98.4 & 99.4 \\
Being-H0.5~\cite{beingh05} & 99.2 & 99.6 & \textbf{99.4} & 97.4 & 98.9 \\
PhysBrain 1.0~\cite{physbrain10} & 99.6 & 99.6 & \textbf{99.4} & 96.4 & 98.8 \\
Xiaomi-Robotics-0~\cite{xiaomirobotics0} & 98.8 & \textbf{100.0} & 98.8 & 97.2 & 98.7 \\
Cosmos Policy~\cite{cosmospolicy} & 98.1 & \textbf{100.0} & 98.2 & 97.6 & 98.5 \\
MolmoAct2-Think~\cite{molmoact2} & 98.8 & 99.8 & 98.5 & 95.4 & 98.1 \\
$\pi_{0.5}$~\cite{pi05} & 98.8 & 98.2 & 98.0 & 92.4 & 96.9 \\
NVIDIA GR00T N1.7~\cite{grootn17} & 97.7 & 97.5 & 98.5 & 94.4 & 97.0 \\
\midrule
\multicolumn{6}{l}{\textit{Closed-loop or online post-training}} \\
Online SRPO~\cite{srpo2025} & 98.8 & \textbf{100.0} & \textbf{99.4} & 98.6 & 99.2 \\
SimpleVLA-RL (OpenVLA-OFT)~\cite{simplevlarl} & 99.4 & 99.1 & 99.2 & 98.5 & 99.1 \\
OpenVLA-OFT (RLinf-GRPO)~\cite{rlinfvla} & 99.40 & 99.80 & 98.79 & 93.95 & 97.99$^{\dagger}$ \\
\textbf{HABILIS Brain 0} & 99.4 & \textbf{100.0} & \textbf{99.4} & \textbf{99.4} & \textbf{99.55} \\
\bottomrule
\end{tabular}
\end{table*}

\subsection{GCRF Evaluation}
\label{sec:gcrf-eval}
With the GC-VLA base parameters fixed, the final GCRF configuration
achieves 99.55\% success on LIBERO, compared with 95.20\% for
GC-VLA alone (Table~\ref{tab:dlf-rollout}). The evaluated configuration
applies the selected bounded velocity correction at all ten flow
integration steps within each replanning cycle.

\subsection{Mechanism Analysis: GC Reader and Residual Placement}
\label{sec:mechanism_ablation}
Having established the full-policy performance above, we next examine two
mechanisms specific to the proposed architecture: whether the GC reader
contributes additional information at inference time, and whether the GCRF
correction benefits from being applied throughout the flow integration
trajectory rather than as a single impulse. These experiments are intended
as targeted mechanism interventions rather than suite-wide performance
comparisons.

We evaluate these interventions on a fixed four-task diagnostic panel:
Spatial/task5, Long/task9, Goal/task3, and Object/task0.
Each condition uses continuous batches of five across 50 initial
states per task, with seed 1000 and ten base flow evaluations.
Table~\ref{tab:mechanism_ablation} reports task-level success rates.

\begin{table*}[t]
\centering
\footnotesize
\setlength{\tabcolsep}{3pt}
\renewcommand{\arraystretch}{1.08}
\caption{Inference-time mechanism interventions. Entries are success
rates (\%) on individual tasks, not suite-wide scores. GCRF is disabled
for the reader interventions. Residual-placement interventions retain
the same learned router and residual and match the integrated
correction within each solve. The panel mean equally weights the four
tasks.}
\label{tab:mechanism_ablation}
\begin{tabular}{lrrrrr}
\toprule
Intervention & Spatial/t5 & Long/t9 & Goal/t3 & Object/t0 & Panel mean \\
\midrule
Full GC reader & 58.0 & 78.0 & 94.0 & 100.0 & \textbf{82.5} \\
Reader removed & 66.0 & 62.0 & 92.0 & 100.0 & 80.0 \\
Head-supervised GC slots & 58.0 & 74.0 & 98.0 & 98.0 & 82.0 \\
Wrist-supervised GC slots & 58.0 & 76.0 & 96.0 & 98.0 & 82.0 \\
\midrule
Residual at all ten steps & 94.0 & 96.0 & 100.0 & 100.0 & \textbf{97.5} \\
Terminal-action impulse & 62.0 & 86.0 & 98.0 & 100.0 & 86.5 \\
First-step impulse & 64.0 & 86.0 & 96.0 & 100.0 & 86.5 \\
\bottomrule
\end{tabular}
\end{table*}

\paragraph{GC reader intervention.}
The full reader achieves 82.5\% compared with 80.0\% when removed,
with task-dependent effects: retaining the reader improves Long/task9
but reduces success on Spatial/task5. Head-only and wrist-only slot
conditions both achieve 82.0\%. None of the three planned reader
contrasts is significant after Holm correction (adjusted $p=1.0$).
Thus, this intervention provides no evidence for a consistent
inference-time benefit from the reader pathway alone. Importantly,
removing the reader does not remove geometry-change pretraining or
the resulting backbone representation; the remaining backbone
conditioning is retained, and the masked slots are contextualized.
The result therefore suggests that the gains of GC-VLA should not be
attributed primarily to the incremental reader pathway, but does not
isolate the contribution of geometry-change pretraining itself.

\paragraph{Residual placement.}
The deployed residual is constant within a replan, while the base
velocity is reevaluated along the evolving flow trajectory.
We compare its application at all ten steps with adding the
accumulated impulse to the terminal action or concentrating it at
the first step. All-step application achieves 97.5\%, versus 86.5\%
for either alternative. Both matched-outcome contrasts yield
Holm-adjusted $p=2.38\times10^{-6}$ across the five planned comparisons.
The results support distributed in-flow application on this panel.
They do not compare independently trained action-residual policies.
Impulse matching does not match peak amplitude or squared control
energy, and later replans can differ across closed-loop trajectories.
Object/task0 has no nonzero residual exposure and therefore supplies
no evidence about placement despite its equal success rates.

\subsection{LIBERO-PRO Four-Axis Perturbation Evaluation}
We evaluate the frozen GC-VLA+GCRF policy on LIBERO-PRO~\cite{liberopro} after
post-training on canonical LIBERO. No LIBERO-PRO-specific training
or parameter updates are performed. Table~\ref{tab:libero-pro}
reports the four selected perturbation axes: language, object,
position, and task. Original and environment conditions are excluded
from this four-axis comparison.

\begin{table}[t]
\centering
\footnotesize
\setlength{\tabcolsep}{3pt}
\renewcommand{\arraystretch}{1.08}
\caption{LIBERO-PRO success rates (\%) over the four reported
perturbation axes. The evaluated GC-VLA+GCRF policy is frozen after
canonical LIBERO post-training. Average denotes the unweighted mean
over the displayed axes. Original and environment conditions are
not included.}

\label{tab:libero-pro}
\begin{tabular}{lccccc}
\toprule
Method
& Language / Semantic
& Object
& Swap / Position
& Task
& Average \\
\midrule
OpenVLA-OFT~\cite{openvlaoft}
& 59.00 & 27.63 & 5.75 & 0.63 & 23.25 \\

MolmoAct~\cite{molmoact2025}
& 85.75 & 76.00 & 1.50 & 1.50 & 41.19 \\

$\pi_{0}$~\cite{pi0}
& 90.50 & 90.50 & 0.00 & 0.00 & 45.25 \\

X-VLA~\cite{xvla2025}
& 82.88 & 78.38 & 1.63 & 16.38 & 44.81 \\

VLA-Adapter~\cite{vlaadapter2025}
& 90.75 & 73.75 & 0.00 & 19.75 & 46.06 \\

SimVLA~\cite{simvla2026}
& \textbf{98.75} & 81.50 & 8.25 & 6.00 & 48.63 \\

$\pi_{0.5}$~\cite{pi05}
& 95.80 & \textbf{96.00} & \textbf{20.80} & 0.80 & \textbf{53.35} \\

\textbf{HABILIS Brain 0}
& 93.00 & 71.85 & 18.05 & \textbf{21.45} & 51.09 \\
\bottomrule
\end{tabular}
\end{table}

GC-VLA+GCRF achieves a 51.09\% macro-average across the four reported
LIBERO-PRO perturbation axes, comparable to the 53.35\% macro-average of
$\pi_{0.5}$. Performance varies substantially by
perturbation type, as is also observed for the comparison methods.
GC-VLA+GCRF obtains 93.00\% on language perturbations, 71.85\% on object
perturbations, 18.05\% on position perturbations, and 21.45\% on task
perturbations. For comparison, $\pi_{0.5}$ obtains 95.80\%, 96.00\%,
20.80\%, and 0.80\% on the same four axes, respectively, with a
53.35\% macro-average. Other methods in Table~\ref{tab:libero-pro}
have macro-averages ranging from 23.25\% to 48.63\%. These results
show different perturbation profiles across policies rather than a
uniform advantage of one method across all axes.

For GC-VLA+GCRF, the LIBERO-PRO evaluation uses the same policy frozen
after closed-loop post-training on canonical LIBERO, without
LIBERO-PRO-specific training or parameter updates. The result therefore
measures how the canonical-LIBERO-trained policy behaves under the
LIBERO-PRO perturbations, rather than the effect of adaptation to those
perturbations. Because LIBERO-PRO is derived from the same underlying
benchmark and task families, we treat this evaluation as a perturbation
robustness test and do not interpret it as evidence of general
embodiment or environment transfer.

\section{Limitations}
GC-VLA and GCRF address different stages of policy learning.
Geometry-change supervision shapes the representation, while the
final 99.55\% LIBERO result additionally uses closed-loop
post-training on the benchmark distribution. The result therefore
characterizes the combined training procedure, rather than
pretraining scale alone. LIBERO-PRO
measures perturbation robustness within related simulated task
families. Its combined-system result does not isolate the causal
contribution of GCRF without a matched base-policy comparison.

\section{Conclusion}

We introduced GC-VLA, which learns action-relevant visual
representations through multiview geometry-change supervision.
Geometry pretraining is followed by detached action alignment and
coupled policy adaptation. The resulting GC-VLA policy achieves
95.20\% success on LIBERO. GCRF augments the frozen policy with a
binary intervention router and one bounded residual velocity policy,
reaching 99.55\% under the reported evaluation protocol.

\bibliographystyle{plain}
\bibliography{references}

\end{document}